\documentclass[conference]{IEEEtran}
\IEEEoverridecommandlockouts
\usepackage{cite}
\usepackage{amsmath,amssymb,amsfonts}
\usepackage{graphicx}
\usepackage{float}
\usepackage[caption=false,font=footnotesize]{subfig}
\usepackage{textcomp}
\usepackage{xcolor}
\usepackage{booktabs}
\usepackage{multirow}
\usepackage{arydshln}
\usepackage{url}
\usepackage{hyperref}
\usepackage{pgfplots}
\pgfplotsset{compat=1.16}
\def\BibTeX{{\rm B\kern-.05em{\sc i\kern-.025em b}\kern-.08em
    T\kern-.1667em\lower.7ex\hbox{E}\kern-.125emX}}

\newcommand\blfootnote[1]{%
  \begingroup
  \renewcommand\thefootnote{}\footnotetext{#1}%
  \addtocounter{footnote}{-1}%
  \endgroup
}

\begin{document}

\title{When Synthetic Speech Is All You Have: \\Better Call GRPO}

\author{
\IEEEauthorblockN{
Shashi Kumar$^{1,2,\star}$,
Yanis Labrak$^{1,\star}$,
Hasindri Watawana$^{1,2}$, 
Sergio Burdisso$^{1}$,  \\
Esaú Villatoro-Tello$^{1}$,
Kadri Hacioglu$^{3}$,
Petr Motlicek$^{1,4}$,
Andreas Stolcke$^{3}$
}
\vspace{3mm}
\IEEEauthorblockA{
$^{1}$\textit{Idiap Research Institute, Martigny, Switzerland};
$^{2}$\textit{EPFL, Lausanne, Switzerland} \\
$^{3}$\textit{Uniphore, U.S.A.};
$^{4}$\textit{Brno University of Technology, Brno, Czech Republic}
}
}

\maketitle

\begin{abstract}
LLM-based ASR adapted to regulated domains such as banking is bottlenecked by privacy: real speech is costly and legally constrained to collect, making synthetic text-to-speech (TTS) an attractive substitute. Yet synthetic speech stays acoustically mismatched with real recordings, and work on this gap has stayed within supervised fine-tuning (SFT). We instead turn to reinforcement learning, and show that Group Relative Policy Optimization (GRPO) extracts far more from the same synthetic speech than SFT. Synthetic-only adaptation of the model with GRPO, a critic-free method rewarding low-WER hypotheses, reduces WER by 40\% relative to SFT (36.71\%$\to$22.09\%), and an SFT-then-GRPO combination pushes this further to 45\%. We trace the gain to behavior rather than representation: GRPO reduces insertion errors by improving stopping calibration and speech-to-text alignment by better anchoring attention to audio, leaving early-layer representations intact. When synthetic speech is the main resource, reinforcement learning should be preferred over supervised fine-tuning.

\end{abstract}


\section{Introduction}

\blfootnote{$^\star$Equal contribution. Code: \href{https://github.com/idiap/better-call-grpo}{github.com/idiap/better-call-grpo}}
 
Speech recordings, such as customer--agent telephone calls in regulated domains such as banking are among the hardest data to collect at scale, because privacy is the central concern and regulations such as the GDPR and the EU AI Act~\cite{eu2024aiact} restrict how such recordings, which may qualify as biometric data, can be stored, shared, and reused for training due to personally identifiable and financial information they carry.

These constraints slow the adoption of speech technologies in high-stakes domains, as systems must be adapted to in-domain speech data, yet sourcing such recordings remains slow, costly, and tightly regulated.

To sidestep this problem, the community has explored substituting real in-domain audio with speech synthesized using text-to-speech (TTS) systems from available transcripts~\cite{rossenbach2020ttsaugmentation,rosenberg2019synth,chen2020datacollection,yang2024versatiletts,perrin2025optsynth}. However, synthetic speech still differs acoustically from real recordings~\cite{srinivasavaradhan2025ttsstate}, and this severe mismatch keeps the resulting performance gain sub-optimal, motivating continuation of work on closing the synthetic-to-real gap~\cite{su2024taskarith,chou2025selfrefining}. A recent work convolves synthetic speech with room impulse responses (RIRs) to reproduce real channel characteristics, matching the real-speech baseline while using only a quarter of the real data~\cite{labrak2026-synth}.

These works typically build on pre-trained large language model (LLM) architectures adapted to the speech modality by plugging in a speech encoder~\cite{tang2024salmonn,chu2023qwenaudio,rubenstein2023audiopalm,ma2024slamasr,ma2024slam,radford2023whisper}, yet they all share the same limitation of remaining within the supervised fine-tuning (SFT) paradigm. Meanwhile, the broader NLP community has moved beyond cross-entropy and successfully leverages synthetic data~\cite{peng2023instructiontuninggpt4,tunstall2024zephyr,padhi-etal-2024-value} through reinforcement learning (RL) approaches such as Direct Preference Optimization (DPO)~\cite{3666122.3668460} and Group Relative Policy Optimization (GRPO)~\cite{shao2024deepseekmath}.

These observations, combined with the acoustic irregularities of synthetic speech, lead us to hypothesize that this reliance on SFT is what caps the performance achievable with synthetic speech. Token-level cross-entropy forces the model to reproduce the reference token by token, so local synthetic artifacts, such as: unnatural prosody, phonetic glitches, or imperfect RIR simulation, get absorbed into fluent but acoustically unsupported continuations, producing hallucinated insertions. RL instead optimizes a sequence-level objective over sampled hypotheses, directly targeting the evaluation metric (e.g.\ WER), so only the output as a whole needs to score well, making it robust to such locally corrupted signal.

For these reasons, this paper focuses on GRPO, a critic-free method that scores a group of sampled hypotheses and reinforces those that outperform the group, using task metrics as the reward function. Its value for adaptation with synthetic speech remains untested, and we hypothesize that it is precisely in this regime that GRPO becomes decisive, making synthetic data usable without relying on real speech. We organize the study around six questions:

\begin{enumerate}
    
    \item \textbf{Is GRPO more effective than SFT?} On real and on synthetic speech, does optimizing WER with GRPO improve over the SFT checkpoint it starts from? (Section~\ref{sec:baselines})

    \item \textbf{Is additional real data necessary?} Given a large synthetic speech corpus, how much real speech is worth annotating, and where do returns diminish? (Section~\ref{sec:realsweep})

    \item \textbf{How much synthetic speech is needed when a small amount of data is available?} With a tiny real speech budget fixed, how do metrics scale with the amount of synthetic speech? (Section~\ref{sec:synthsweep})
    
    \item \textbf{Which reward function is more optimal?} Among WER, CER, a length term, and their combinations, which reward function minimizes errors? (Section~\ref{sec:reward})
    
    \item \textbf{Does selecting a subset of the synthetic training speech help?} Do entropy-based selection strategies over speaker, semantic, acoustic embeddings, or GRPO-specific selectors such as ranking utterances by rollout statistics, outperform random sampling? (Section~\ref{sec:filtering})

    \item \textbf{Why does GRPO work better than SFT?} Is its advantage behavioral rather than representational? How is the attention anchored to audio inputs after different adaptations approaches? (Section~\ref{sec:why})
 
\end{enumerate}

\section{Method}

In this section, we first describe the LLM-based ASR used throughout our experiments and motivate why this architecture provides a controlled setting for studying domain adaptation through post-training. We then describe supervised fine-tuning (SFT), the predominant adaptation approach. Next, we discuss why synthetic speech adaptation may benefit from a sequence-level objective, and describe our application of Group Relative Policy Optimization (GRPO)~\cite{shao2024deepseekmath} to ASR. Finally, we define the reward functions used during GRPO training.

\subsection{LLM-based ASR Backbone}

We use a modular LLM-based ASR backbone that connects a WavLM-Large speech encoder~\cite{chen2022wavlm} to Llama-3.2-1B-Instruct~\cite{dubey2024llama3} through a trainable speech projector. Following the SLAM-ASR recipe~\cite{ma2024slam,ma2024slamasr}, the projector consists of two linear layers with a ReLU activation and downsamples the acoustic sequence by a factor of $5$ before passing it to the LLM. During adaptation, we update the projector and LoRA adapters~\cite{hu2022lora} inserted into the LLM. The LoRA rank is set to $r=16$ with scaling factor $\alpha=32$. The WavLM encoder and the LLM parameters remain frozen throughout all experiments.

We choose this backbone to provide a controlled setting for studying domain-adaptive post-training. Speech-capable LLMs, such as Phi-4-Multimodal~\cite{abouelenin2025phi} and Qwen3-ASR~\cite{shi2026qwen3}, have undergone large-scale speech or multimodal post-training, often including synthetic speech. This makes it difficult to isolate the effects of a downstream adaptation objective and its data. In contrast, our backbone combines a frozen speech encoder with a frozen text LLM, restricting adaptation to the projector and LoRA parameters. This design is suited for evaluating how synthetic speech adaptation data and the training objective affect ASR performance.

\subsection{Domain Adaptation using Supervised Fine-Tuning}
\label{sec:sft-method}

Supervised fine-tuning (SFT) is the predominant approach for adapting LLM-based ASR models~\cite{burdisso2026_TOA,baneras2026closing,villatoro_2026_context_proj,burdisso2025_prompt_projection}. Given an input utterance $x$ and its reference transcript $y^\star = (y^\star_1,\dots,y^\star_T)$, SFT trains the model with teacher forcing by minimizing the token-level cross-entropy loss:
\begin{equation}
\mathcal{L}_{\mathrm{SFT}}(\theta)
=
-\mathbb{E}_{(x,y^\star)}
\left[
\sum_{t=1}^{T}
\log \pi_\theta
\left(
y^\star_t \mid x, y^\star_{<t}
\right)
\right].
\end{equation}
This objective optimizes the likelihood of each reference token under the ground-truth prefix. However, when the adaptation data includes synthetic speech, the acoustic signal can contain prosodic mismatch, phonetic artifacts, or imperfect articulation. Since SFT applies a local token-level penalty at every position, it can be sensitive to such localized acoustic artifacts and may overfit to token-level mismatches.

\subsection{Domain Adaptation using GRPO}
\label{sec:grpo-method}

Unlike SFT, GRPO optimizes a sequence-level objective over model-generated hypotheses, shifting adaptation from matching every reference token under the ground-truth prefix to comparing complete decoded transcripts under the model's own generation distribution. Because rewards are assigned to entire transcripts rather than individual tokens, localized artifacts in synthetic speech should have less influence on the training signal. In this sense, GRPO provides a discriminative sequence-level adaptation signal: rather than only increasing the likelihood of the reference transcript as in SFT, it reinforces good hypotheses and suppresses lower-reward alternatives for the same utterance.
This contrastive effect may help reduce overfitting to synthetic-speech artifacts, since the update is driven by separation between better and worse transcripts rather than token-level imitation alone.
We therefore study GRPO for synthetic speech domain adaptation, following its recent application to ASR on real speech~\cite{11434657}.

Let $x$ be an input utterance, real or synthetic, with reference transcript $y^\star$. For each $x$, the old policy $\pi_{\theta_{\mathrm{old}}}$ samples a group of $G$ hypotheses $\{o_1,\dots,o_G\}$ by stochastic decoding with temperature $\tau$. Each hypothesis receives a scalar reward $r_i = R(o_i, y^\star)$, where $R$ is one of the reward functions defined in Section~\ref{sec:rewards}. Sampling a group of hypotheses exposes the model to alternative transcripts for the same utterance, allowing the update to compare relatively better and worse decodes under the same acoustic input.
GRPO computes a group-relative advantage by standardizing the sampled rewards:
\begin{equation}
  A_i \;=\; \frac{r_i - \operatorname{mean}(\{r_1,\dots,r_G\})}{\operatorname{std}(\{r_1,\dots,r_G\}) + \varepsilon_{\text{std}}},
\end{equation}
This normalization removes the need for a learned value function and makes the update depend on relative transcript quality within each sampled group, reducing the need to calibrate reward scales across utterances.
The policy is updated by maximizing the clipped surrogate objective with a KL penalty to a fixed reference policy $\pi_\text{ref}$ (the SFT checkpoint):
\begin{equation}
\begin{split}
  \mathcal{J}(\theta) = \;& \mathbb{E}\Big[\tfrac{1}{G}\textstyle\sum_{i=1}^{G}\tfrac{1}{|o_i|}\textstyle\sum_{t}
   \min\!\big(\rho_{i,t}A_i,\; \\
  & \operatorname{clip}(\rho_{i,t},1{-}\epsilon,1{+}\epsilon)A_i\big)\Big]
   - \beta  \mathrm{KL}\!\left[\pi_\theta  \|  \pi_\text{ref}\right],
\end{split}
\end{equation}
where $\rho_{i,t} = \pi_\theta(o_{i,t}\mid x, o_{i,<t}) / \pi_{\theta_\text{old}}(o_{i,t}\mid x, o_{i,<t})$ is the per-token importance ratio, $\epsilon$ is the clipping range, and $\beta$ is the KL coefficient. The reference policy is fixed during GRPO training. Hypotheses with above-average reward receive positive advantage and are reinforced, while lower-reward hypotheses are suppressed. The KL term limits drift from the reference policy, preventing the model from exploiting the reward at the cost of transcription fluency or decoding stability.

\subsection{Reward Functions for GRPO}
\label{sec:rewards}
We define rewards so that higher values indicate better recognition. Our default reward is \textsc{Wer}, defined as one minus standard word error rate (WER), $R_\text{WER}(o,y^\star) = 1-\operatorname{WER}(o,y^\star)$, which directly targets the evaluation metric. We additionally study a character-level reward $R_\text{CER}= 1-\operatorname{CER}$, a length-consistency term $R_\text{LEN}$ that penalizes the length normalized absolute difference in token count between hypothesis and reference, and the additive combinations \textsc{Wer} + \textsc{Len}, \textsc{Cer} + \textsc{Len}, \textsc{Wer} + \textsc{Cer}, and \textsc{Wer} + \textsc{Cer} + \textsc{Len}. All reward experiments are otherwise identical.

\section{Experimental Setup}

%
\subsection{Data}
All experiments use DefinedAI~\cite{kumar2025distilling, burdisso2026_TOA}, a corpus of manually transcribed English customer--agent telephone calls, restricted to its banking domain ($\simeq$54 h of real speech for training, 6.55 h of test set). As a high-resource out-of-domain pre-training source we also use LibriSpeech 960 h \cite{7178964}, named as LS 960 in the rest of the sections. Synthetic counterparts of the banking references transcripts are generated with Qwen3-TTS~\cite{qwen3tts2025}, using VoiceDesign letting us match synthetic voices to real data persona metadata (gender, race) without any reference recordings. The prompting strategy (\emph{``clear articulation, naturally''}) is used to improve prosody as shown by \cite{labrak2026-synth} and convolved with measured RIRs from the BUT Speech@FIT Reverb Database~\cite{szoke2019but} to reproduce the channel acoustics closer to the real recordings. The overall pipeline is following \cite{labrak2026-synth} protocol, using SDialog~\cite{burdisso-etal-2026-sdialog}. We refer to this RIR-augmented synthetic speech simply as synthetic speech throughout. 

\subsection{Training Configurations}
We evaluate two regimes. The first assumes access to the full 54 h of real banking speech and is used to compare SFT and GRPO and to study adaptation from an LS 960 checkpoint. This experiment represent the upper limit of what is possible with a high budget. The second assumes only a tiny budget (as little as 1 h of real speech) alongside the synthetic speech pool, reflecting the privacy-constrained setting.

\subsection{Hyperparameters}
Both SFT and GRPO uses AdamW, learning rate $1\times10^{-4}$ and $2\times10^{-5}$ respectively, batch size $16$, $100$ warmup steps, $5$ epochs, with LoRA $r=16$, $\alpha=32$. GRPO starts from SFT checkpoints, rollout group size $G=4$, maximum $128$ new tokens, sampling temperature $\tau=0.8$, KL coefficient $\beta=0.04$, clipping $\epsilon=0.2$, and gradient-norm clipping at $1.0$. All runs use BF16 precision on a single NVIDIA H100 80 GB GPU. During inference, the models are allowed to generate maximum $256$ new tokens, with beam size $1$.

\subsection{Metrics}
We report word error rate (WER), character error rate (CER), as well as insertion (INS), deletion (DEL), and substitution (SUB) rates. Lower is better for all metrics.

\subsection{Synthetic Speech Selection}
\label{sec:selection}

When the synthetic pool is larger than the training budget, the choice of which utterances to keep may affect how much useful signal GRPO receives. We compare six selection strategies. The first is random sampling, which serves as the main baseline. The next three are SFT-style diversity selectors: each utterance is embedded\footnote{\href{https://huggingface.co/pyannote/embedding}{huggingface.co/pyannote/embedding}} using either a speaker encoder (pyannote~\cite{bredin2023pyannote}), a semantic text encoder (BGE~\cite{xiao2024bge}), or the acoustic encoder (WavLM~\cite{chen2022wavlm}), and a greedy maximum-entropy procedure selects utterances that maximize coverage in that embedding space.

The final two selectors are designed for GRPO: we sample $G$ hypotheses per utterance from a fixed checkpoint (as in GRPO training) and rank utterances by rollout statistics. \textsc{Grpo-Signal} favors high reward variance, large reward range, and diverse hypotheses. \textsc{Grpo-Recoverable} favors a large gap between the mean and best sampled hypothesis, i.e.\ cases the model transcribes well only sometimes. For every strategy we construct nested 1 h, 5 h, and 25 h subsets.

\section{Results}

We compare GRPO against SFT and discuss the choices that govern how well it exploits synthetic speech. Models are adapted from the LS 960 SFT checkpoint, unless specified.

\subsection{Does GRPO outperform SFT?}
\label{sec:baselines}

\begin{table}[H]
\centering
\caption{Results on the test set, comparing SFT and GRPO fine-tuning from the LS-960h SFT baseline (\textsc{Base}), using either 54h of real or synthetic banking speech.}
\label{tab:baselines}

\resizebox{\columnwidth}{!}{%
\begin{tabular}{lccl}

\toprule

\textbf{Configuration} & \textbf{WER} $\downarrow$ & \textbf{CER} $\downarrow$ & \textbf{INS/DEL/SUB} \\

\midrule

LS 960 \textsc{Sft} (\textsc{Base}) & 66.44 & 46.73 & 42.18 / 5.68 / 18.59 \\

\midrule

\multicolumn{4}{c}{\textsc{Real Banking Speech}} \vspace{1mm} \\

$\rightarrow$ \textsc{Sft}                           & 10.27 & 7.11  & 2.99 / 2.13 / 5.14  \vspace{0.5mm} \\

$\rightarrow$ \textsc{Grpo}                         & 10.78 & 7.68  & 1.79 / 3.50 / 5.49 \vspace{0.5mm} \\

$\rightarrow$ \textsc{Sft} $\rightarrow$ \textsc{Grpo}    & \textbf{9.49} & \textbf{6.66} & 1.98 / 2.50 / 5.01 \\

\midrule

\multicolumn{4}{c}{\textsc{Synthetic Banking Speech}} \vspace{1mm} \\
$\rightarrow$ \textsc{Sft}                                & 36.71 & 25.06 & 24.79 / 2.55 / 9.37  \vspace{0.5mm} \\
$\rightarrow$ \textsc{Grpo}                               & 22.09 & 15.89 & 8.39 / 5.15 / 8.56   \vspace{0.5mm} \\
$\rightarrow$  \textsc{Sft} $\rightarrow$ \textsc{Grpo}                 & \textbf{20.21} & \textbf{14.68} & 8.60 / 3.66 / 7.95 \\
\bottomrule
\end{tabular}%
}
\label{table:main}
\end{table}

Two observations stand out in Table~\ref{tab:baselines}. First, on real data, GRPO helps mainly when stacked after SFT. Applied directly from LS 960, GRPO does not outperform SFT on real speech alone, but SFT followed by GRPO extracts a small additional gain even where abundant real data exists, driven primarily by reduced insertions.

Second, the effect is far larger when only synthetic speech is available. SFT on synthetic speech alone performs poorly (36.71\% WER), dominated by insertions (24.79), whereas GRPO over the same data reaches 22.09\% and SFT followed by GRPO 20.21\%, a 16 point absolute reduction over SFT, or 45\% relative. GRPO thus recovers a large fraction of the synthetic-to-real gap precisely in the privacy-constrained setting that motivates synthetic speech in the first place, validating our initial intuition.
\subsection{How Much Real Data Is Worth Collecting?}
\label{sec:realsweep}

We first ask how much real speech is worth pairing with the synthetic pool.
Starting from LS 960 SFT, we fix the 54 h synthetic pool and add increasing amounts of real banking speech, all optimized with GRPO (Table~\ref{tab:realsweep}).

\begin{table}[H]
\centering
\caption{Effect of varying the amount of real banking speech ($X$ h) mixed with a fixed 54\,h of synthetic speech, fine-tuned with GRPO on top of the LS 960 SFT baseline.}
\label{tab:realsweep}
\begin{tabular}{rcccl}
\toprule
\textbf{Real} & \textbf{Synth} & \textbf{WER} $\downarrow$ & \textbf{CER} $\downarrow$ & \textbf{INS/DEL/SUB} \\

\midrule

\textsc{0 h}  & \textsc{54 h} & 22.09 & 15.89 & 8.39 / 5.15 / 8.56 \\

\midrule

\textsc{1 h}  & \multirow{5}{*}{\textsc{54 h}} & 20.32 & 14.08 & 7.25 / 4.16 / 8.91 \\
\textsc{5 h}  &                     & 15.51 & 11.41 & 3.35 / 4.65 / 7.51 \\
\textsc{10 h} &                     & 14.03 & 10.10 & 2.90 / 3.89 / 7.24 \\
\textsc{25 h} &                     & 13.03 & 9.59  & 3.10 / 3.36 / 6.57 \\
\textsc{54 h} &                     & \textbf{12.61} & \textbf{9.03}  & 3.58 / 3.20 / 5.83 \\
\bottomrule
\end{tabular}
\end{table}

WER falls steeply over the first few hours of real data, from 20.32\% at 1h to 15.51\% at 5h and 14.03\% at 10h, then flattens, improving only to 13.03\% at 25h and 12.61\% at the full 54h.
The marginal value of real speech is thus concentrated in the first $\approx$5--10 hours: roughly two thirds of the achievable WER reduction comes from less than a fifth of the real data, after which returns diminish relative to collection and retention cost.

\subsection{Do We Need All The Synthetic Speech?}
\label{sec:synthsweep}

The complementary question holds the real budget at its minimum (1 h) and instead varies the amount of synthetic speech used for GRPO (Table~\ref{tab:synthsweep}).

\begin{table}[H]
\centering
\caption{Effect of varying the amount of synthetic speech ($X$ h) mixed with a fixed 1\,h of real in-domain speech, with GRPO on top of the LS 960 SFT baseline.}
\label{tab:synthsweep}
\begin{tabular}{rcccl}
\toprule
\textbf{Synth} & \textbf{Real} & \textbf{WER} $\downarrow$ & \textbf{CER} $\downarrow$ & \textbf{INS/DEL/SUB} \\
\midrule

\textsc{25 h} & \textsc{0 h} & 22.11 & 15.62 & 8.62 / 4.04 / 9.44 \\

\textsc{54 h}  & \textsc{0 h} & 22.09 & 15.89 & 8.39 / 5.15 / 8.56 \\
\midrule
\textsc{1 h}  & \multirow{5}{*}{\textsc{1 h}} & 24.60 & 17.59 & 6.60 / 6.55 / 11.45 \\
\textsc{5 h}  &                    & 24.20 & 16.49 & 8.45 / 5.40 / 10.34 \\
\textsc{10 h} &                    & 19.91 & 13.77 & 5.22 / 5.27 / 9.42  \\
\textsc{25 h} &                    & \textbf{16.77} & \textbf{11.63} & 3.82 / 4.46 / 8.48 \\
\textsc{54 h} &                    & 20.32 & 14.08 & 7.25 / 4.16 / 8.91 \\
\midrule
\textsc{25 h} &     \textsc{5 h}               & 14.56 & 10.50 & 3.10 / 4.25 / 7.21 \\
\bottomrule
\end{tabular}
\end{table}

The relationship is non-monotonic. With 1 h of real speech fixed, WER improves as synthetic speech grows from 1 h (24.60\%) to 25 h (16.77\%), but degrades when all 54 h are used (20.32\%). Thus, more synthetic speech is not automatically better: after a moderate point, the additional synthetic data appears to weaken the influence of the limited real in-domain signal rather than provide useful diversity. The best 1 h setting uses roughly 25 h of synthetic speech, giving the strongest balance between synthetic coverage and real-speech anchoring. Increasing the real budget from 1 h to 5 h at the same 25 h synthetic setting improves WER further to 14.56\%, showing that GRPO continues to benefit from additional real supervision once the synthetic amount is kept in a favorable range. We therefore adopt the 25 h synthetic setting as the starting point for the reward study that follows.

\subsection{Which Reward Function Is Best?}
\label{sec:reward}

Holding this data mix fixed, we vary only the GRPO reward (Table~\ref{tab:reward}) as mentioned in the Section \ref{sec:rewards}.

The \textsc{Wer} reward yields the best \%WER (16.77), confirming the principle of optimizing the target metric directly. Adding a \textsc{Len} or \textsc{Cer} term does not help and often hurts, \textsc{Wer+Len}, \textsc{Wer+Cer}, and the three-way combination all trail it, and the extra terms tend to raise insertions. \textsc{Cer} behave as expected on their own metric, with \textsc{Cer+Len} achieving the best \%CER (11.02) at a competitive \%WER (16.81), a reasonable choice when character-level accuracy is the priority. For \%WER-driven evaluation, the \textsc{Wer} is thus both the simplest and the strongest option. With the objective and data quantities settled, one lever remains, which synthetic utterances to train on.

\begin{table}[H]
\centering
\caption{Ablation of the GRPO reward function, fine-tuned on 25\,h synthetic + 1\,h real in-domain speech. \textsc{Len} denotes an added length-matching penalty.}
\label{tab:reward}
\begin{tabular}{lcccl}
\toprule
\textbf{Reward Function} & \textbf{WER} $\downarrow$ & \textbf{CER} $\downarrow$ & \textbf{INS/DEL/SUB} \\
\midrule
\textsc{Wer}             & \textbf{16.77} & 11.63 & 3.82 / 4.46 / 8.48 \\
\textsc{Wer} + \textsc{Len}     & 17.42 & 12.47 & 4.52 / 4.48 / 8.42 \\
\midrule
\textsc{Cer}             & 18.46 & 11.97 & 5.97 / 3.50 / 8.98 \\
\textsc{Cer} + \textsc{Len}     & 16.81 & \textbf{11.02} & 4.22 / 3.67 / 8.91 \\
\midrule
\textsc{Wer} + \textsc{Cer}     & 18.27 & 12.63 & 5.63 / 4.11 / 8.53 \\
\textsc{Wer} + \textsc{Cer} + \textsc{Len} & 17.32 & 11.68 & 4.69 / 3.89 / 8.74 \\
\bottomrule
\end{tabular}
\end{table}

\subsection{Does Selecting Synthetic Speech Help?}
\label{sec:filtering}

The last lever is which synthetic utterances to train on. 
In the Table~\ref{tab:filtering}, we compared random sampling, three SFT style diversity selectors (semantic and acoustics embeddings), and the two GRPO-specific selectors described in Section~\ref{sec:selection}.
Every strategy selects nested 1 h, 5 h, and 25 h subsets from the same 54 h synthetic pool, and each subset is then fine-tuned with GRPO from the LS 960 SFT checkpoint. No real speech is considered here.
The result is mixed and budget dependent. At the smallest budgets selection can help, but the best criterion shifts, BGE is best at 1 h (38.65\% vs.\ 40.21\% for random), while speaker diversity selection is best at 5 h (30.50\%), but the effect is unstable at different budgets. On the other hand, WavLM selection is far worse at 1 h (59.67\%), selecting atypical utterances that make the training difficult when the budget is too small.

At 25 h, random sampling is the strongest overall choice. It reaches 22.11\% WER, while speaker diversity and \textsc{Grpo-Recoverable} are close but no better (22.30\% and 22.31\%), and the explicitly GRPO-oriented \textsc{Grpo-Signal} criterion is worse (25.45\%). Rollout-derived selection therefore does not improve over random sampling, even though it directly uses the GRPO hypothesis distribution.

Taken together, GRPO recovers most of the synthetic-to-real gap from a few hours of real speech, a WER reward, and a randomly sampled synthetic pool. What remains open is what GRPO changes in the model, which the next section addresses.

\begin{table}[H]
\centering
\caption{Comparison of synthetic speech selection strategies for GRPO on top of LS 960 SFT, at three subset sizes drawn from the 54\,h synthetic pool. Non-random strategies use greedy maximum-entropy selection in their respective embedding space.}
\label{tab:filtering}
\resizebox{\columnwidth}{!}{%
\begin{tabular}{lrccl}
\toprule
\textbf{Strategy} & \textbf{Synth} & \textbf{WER} $\downarrow$ & \textbf{CER} $\downarrow$ & \textbf{INS/DEL/SUB} \\
\midrule
\multirow{3}{*}{\textsc{Random Selection}}
        & \textsc{1 h}  & 40.21 & 26.50 & 20.03 / 5.47 / 14.71 \\
      & \textsc{5 h}  & 33.82 & 21.30 & 16.27 / 5.07 / 12.48 \\
      & \textsc{25 h} & \textbf{22.11} & 15.62 & 8.62 / 4.04 / 9.44 \\
\midrule
\midrule
\multirow{3}{*}{\textsc{Speaker Embeddings}}
        & \textsc{1 h}  & 40.38 & 26.90 & 20.29 / 5.26 / 14.83 \\
      & \textsc{5 h}  & 30.50 & 21.46 & 13.77 / 4.60 / 12.13 \\
      & \textsc{25 h} & 22.30 & \textbf{15.21} & 9.15 / 3.73 / 9.42 \\
\midrule
\multirow{3}{*}{\textsc{WavLM Embeddings}}
        & \textsc{1 h}  & 59.67 & 44.56 & 38.53 / 5.35 / 15.79 \\
      & \textsc{5 h}  & 32.67 & 22.61 & 15.36 / 4.50 / 12.81 \\
      & \textsc{25 h} & 23.87 & 16.67 & 9.95 / 4.00 / 9.92 \\
\midrule
\midrule
\multirow{3}{*}{\textsc{BGE Embeddings}}
        & \textsc{1 h}  & 38.65 & 27.50 & 18.30 / 5.16 / 15.20 \\
      & \textsc{5 h}  & 35.60 & 23.82 & 17.73 / 4.63 / 13.24 \\
      & \textsc{25 h} & 25.01 & 16.65 & 11.49 / 3.17 / 10.35 \\
\midrule
\midrule
\multirow{3}{*}{\textsc{GRPO-Recoverable}}
        & \textsc{1 h}  & 43.20 & 30.05 & 23.14 / 5.60 / 14.47 \\
      & \textsc{5 h}  & 33.85 & 24.96 & 17.53 / 4.21 / 12.11 \\
      & \textsc{25 h} & 22.31 & 15.34 & 9.19 / 3.45 / 9.67 \\
\midrule
\multirow{3}{*}{\textsc{GRPO-Signal}}
        & \textsc{1 h}  & 42.22 & 28.94 & 22.34 / 5.72 / 14.16 \\
      & \textsc{5 h}  & 35.28 & 23.82 & 18.31 / 4.42 / 12.54 \\
      & \textsc{25 h} & 25.45 & 18.31 & 12.24 / 3.15 / 10.06 \\
\bottomrule
\end{tabular}%
}
\end{table}

\section{Why Does GRPO Work Better than SFT?}
\label{sec:why}

When only synthetic speech is available, SFT fails mainly through insertions.
In Table~\ref{tab:baselines}, synthetic SFT reaches 36.71\% WER, of which
24.79 points are insertions. GRPO reduces WER to 22.09\% and cuts insertions
to 8.39. We interpret this as a difference in how the two objectives handle
localized synthetic artifacts. Synthetic speech may contain unreliable
regions caused by TTS prosody, phonetic artifacts, or imperfect RIR simulation.
Token-level SFT forces the model to explain every local region against the
reference, so small acoustic mismatch can push the decoder into a
language-model-driven continuation without sufficient acoustics grounding. GRPO instead scores complete hypotheses:
outputs where a local artifact grows into a long insertion chain receive low
reward.

The following analyses trace this mechanism from the output errors, to decoder confidence, to audio-token attention, and finally to representation change.

\subsection{GRPO removes length-independent hallucination}

\begin{figure}[H]
\centering
\includegraphics[width=\columnwidth]{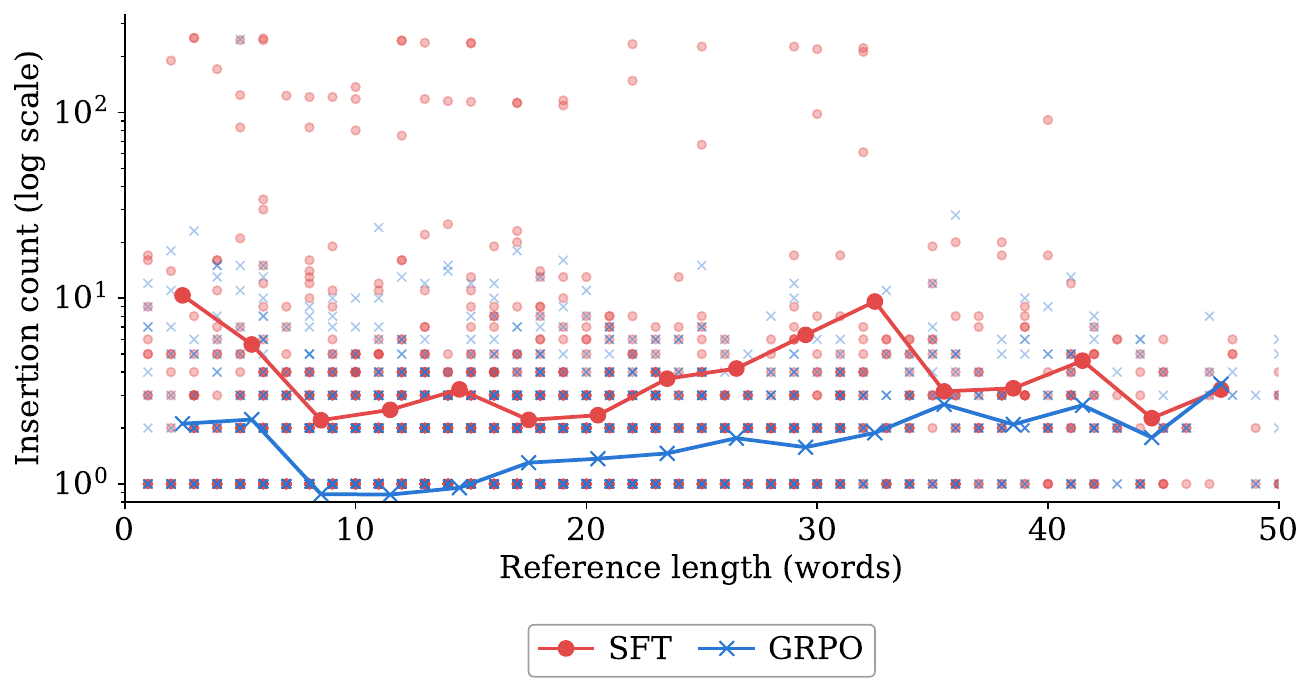}
\caption{Insertion count vs.\ reference length for SFT and GRPO, with individual utterances shown as scattered points.}
\label{fig:inslen}
\end{figure}

We first ask whether the WER gain comes from containing insertion errors.
Figure~\ref{fig:inslen} plots per-utterance insertion count against reference
length. \textsc{Sft} has a heavy insertion tail, including short utterances
followed by tens or even hundreds of extra tokens. This shows that the synthetic
mismatch is not only producing local word errors, it can trigger long
continuations after the spoken content has ended. \textsc{Grpo} largely removes
this tail and keeps insertion counts low across reference lengths. Thus, GRPO
does not merely improve average word recognition, but prevents local synthetic
defects from propagating into global decoding failures.

\subsection{GRPO improves stopping calibration}

\begin{table}[H]
\centering
\caption{Expected Calibration Error (ECE) by utterance length bin (number of reference tokens).}
\label{tab:ece-length}
\begin{tabular}{lcccc}
\toprule
 & \textbf{Short} & \textbf{Medium} & \textbf{Long} & \textbf{Very long} \\
\textbf{Model} & \textbf{\textit{(1--8)}} & \textbf{\textit{(9--15)}} & \textbf{\textit{(16--22)}} & \textbf{\textit{($\geq$23)}} \\
\midrule
\textsc{Base}           & 0.619 & 0.409 & 0.331 & 0.271 \\
\midrule
$\rightarrow$ \textsc{Sft}      & 0.430 & 0.195 & 0.135 & 0.174 \\
$\rightarrow$ \textsc{Grpo}     & 0.198 & \textbf{0.085} & \textbf{0.078} & \textbf{0.080} \\
$\rightarrow$ \textsc{Sft} $\rightarrow$ \textsc{Grpo} & \textbf{0.186} & 0.107 & 0.099 & 0.095 \\
\bottomrule
\end{tabular}%
\end{table}

The insertion tail suggests a stopping failure, after the acoustically supported
transcript ends, the model should become uncertain rather than continue
generating text. We therefore examine token confidence around the final
reference token.

\begin{figure}[H]
\centering
\includegraphics[width=\columnwidth]{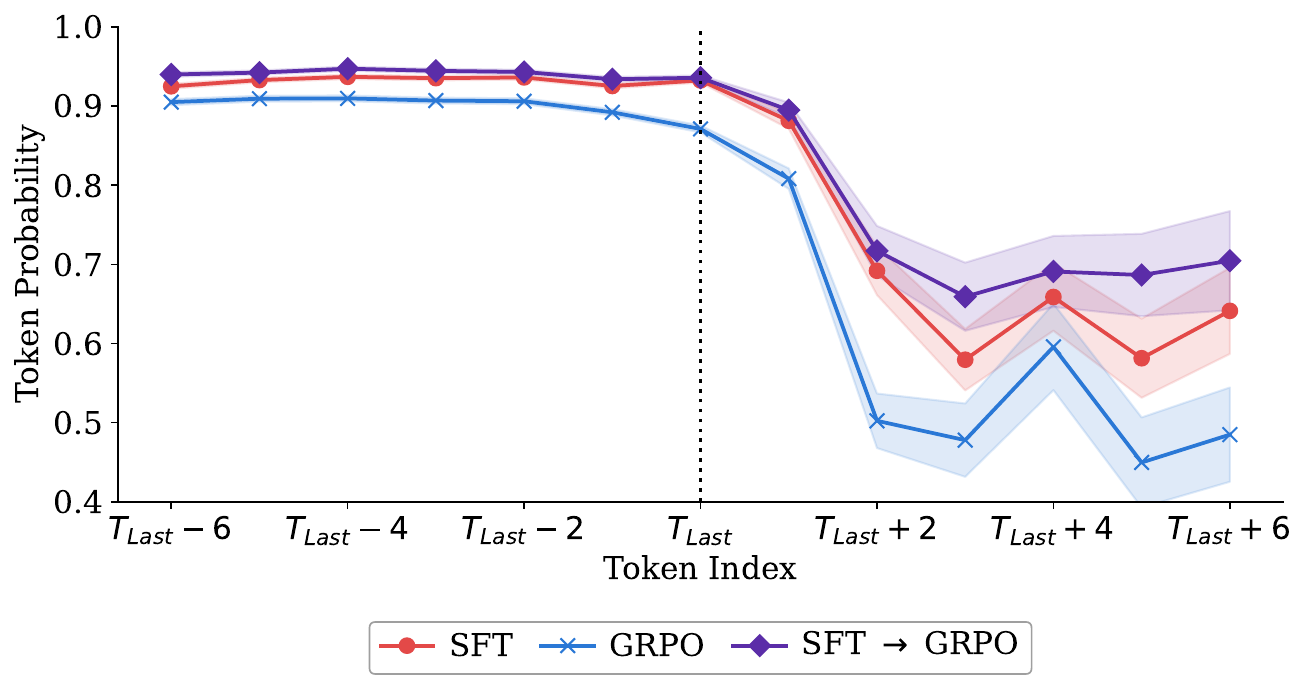}
\caption{Mean token probability (with confidence bands) around the final reference token ($T_{Last}$, dotted line) for SFT, GRPO, and SFT $\rightarrow$ GRPO.}
\label{fig:boundary}
\end{figure}

Figure~\ref{fig:boundary} shows that \textsc{Sft} remains
confident after the boundary, consistent with confident hallucinated
continuation. \textsc{Grpo} drops much more sharply after the final token,
indicating that the policy has learned when not to continue. Table~\ref{tab:ece-length}
confirms this in aggregate, GRPO lowers ECE in every length bin, with the
largest gain on short utterances, where insertion bursts are most damaging.
This supports the view that GRPO turns uncertain local regions into conservative
stopping behavior rather than fluent over-generation.

\subsection{GRPO relies on stable audio evidence}

\begin{figure}[H]
\centering
\includegraphics[width=\columnwidth]{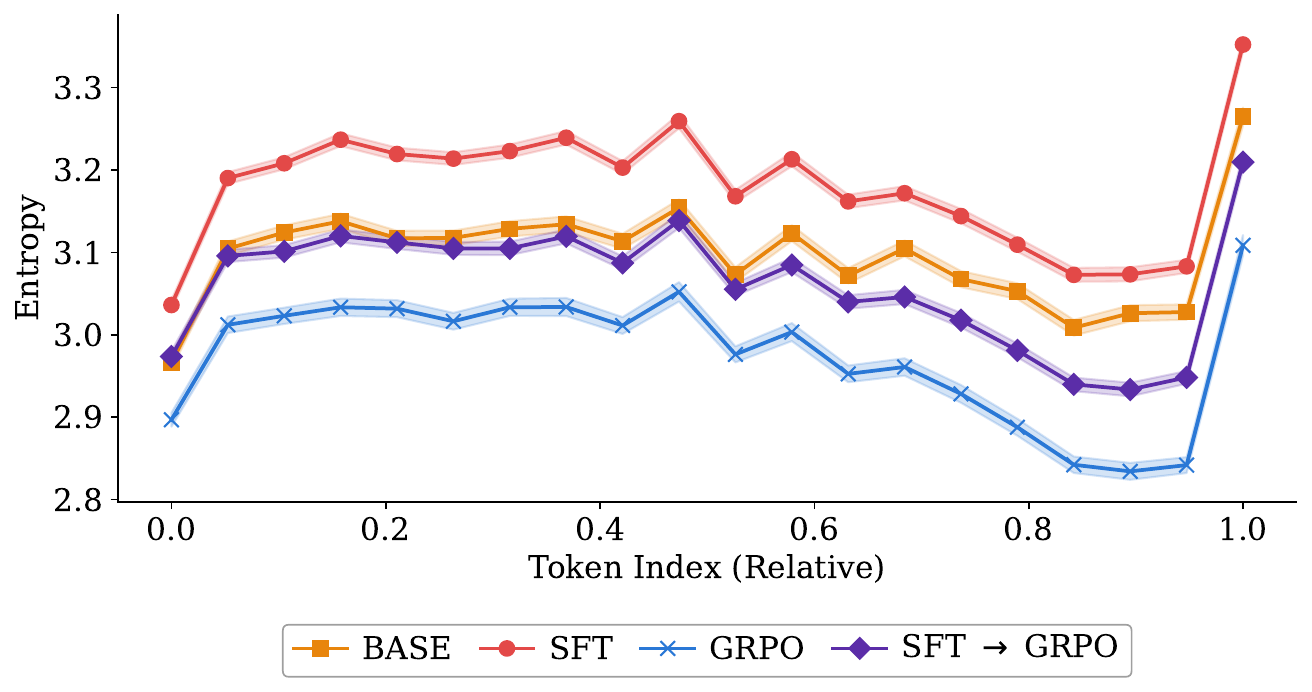}
\caption{Token-level attention entropy over relative decoding position. Higher values indicate attention spread more diffusely over audio tokens; lower values, sharper focus.}
\label{fig:entropy_attn_time}
\end{figure}

If the model is driven mainly by the language prior, attention over audio tokens should become diffuse. Figure~\ref{fig:entropy_attn_time} plots attention entropy over decoding time. \textsc{Sft} has consistently higher entropy than \textsc{Grpo}, indicating weaker concentration on specific audio regions. \textsc{Grpo} keeps attention entropy lower through most of the decoding trajectory, suggesting that it emits tokens only when the audio evidence is stable enough.

\begin{figure}[H]
\centering
\includegraphics[width=0.94\columnwidth]{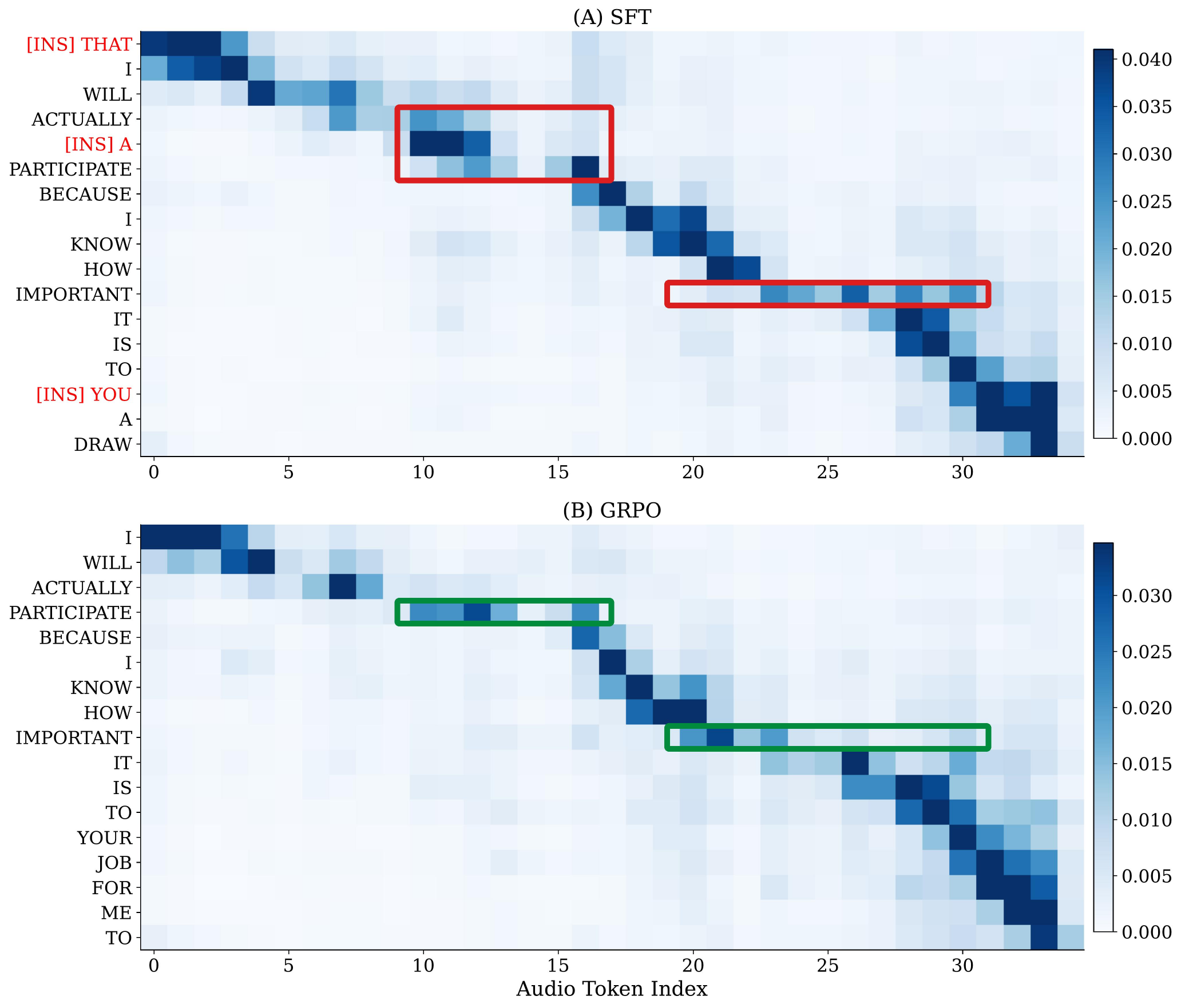}
\caption{Decoder cross-attention for one utterance under \textsc{Sft} (top) and \textsc{Grpo} (bottom); inserted tokens marked \texttt{[INS]} in red. Boxes contrast the same audio regions: \textsc{Sft}'s attention is diffuse and leaks across words (red), while \textsc{Grpo}'s stays sharp on a single row (green).}
\label{fig:attn}
\end{figure}

The heatmap in Figure \ref{fig:attn} shows the same behavior.
\textsc{Sft} emits inserted tokens with weakly localized attention,
while \textsc{Grpo} keeps a cleaner monotonic alignment and produces no inserted
tokens for the same utterance. Together, the entropy and heatmap analyses show
that GRPO makes the decoder more robust to local synthetic artifacts by
discouraging tokens that are not supported by stable audio evidence.

\subsection{GRPO is a lightweight policy correction}

\begin{figure}[H]
\centering
\includegraphics[width=\columnwidth]{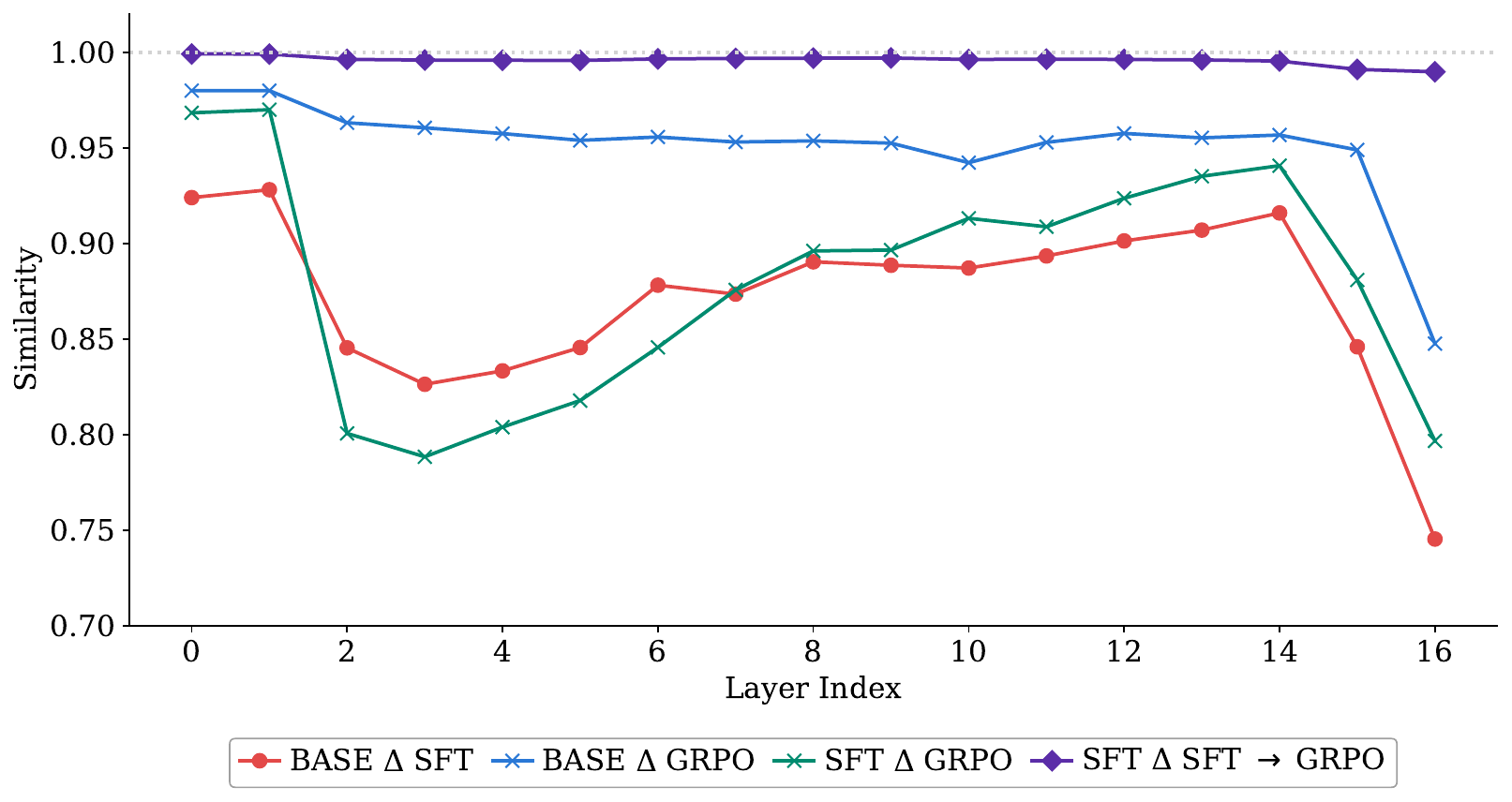}
\caption{Layerwise linear CKA similarity between model pairs. Lower similarity reflect bigger delta in weights.}
\label{fig:cka}
\end{figure}

Finally, we ask whether this robustness comes from broadly rewriting the audio
representations. Figure~\ref{fig:cka} compares layerwise representations using
linear Centered Kernel Alignment (CKA) similarity. \textsc{Grpo} stays close to \textsc{Base} throughout, while
\textsc{Sft} diverges most in the early-to-middle layers, the same region
associated with the synthetic/real gap in previous work~\cite{labrak2026-synth}.
Moreover, \textsc{Sft} $\rightarrow$ \textsc{Grpo} remains almost identical to
\textsc{Sft}, showing that the final GRPO stage changes behavior without
substantially changing the representation geometry.

This completes the mechanism suggested by the previous plots. SFT strongly
adapts the representation to synthetic speech, but can turn local acoustic
artifacts into confident insertion chains. GRPO instead acts as a targeted
policy correction: it preserves most representations, improves stopping
calibration, and suppresses continuations that are not supported by stable
audio evidence.

\section{Conclusion}

We asked whether synthetic speech is better exploited by SFT or by GRPO when privacy and cost rule out collecting real recordings. On a banking domain ASR task, GRPO is clearly the better choice: trained on synthetic speech alone, it cuts WER from 36.71\% (SFT) to 22.09\%, an SFT-then-GRPO schedule reaches 20.21\%, and GRPO adds a small but consistent gain on real data (10.27\% $\to$ 9.49\%).
The gain is behavioral, not representational. Synthetic-only SFT hallucinates confident insertions once acoustic support runs out. GRPO instead improves stopping calibration and keeps attention anchored to stable audio evidence, while leaving the early-layer representations that SFT rewrites largely intact.
Three takeaways follow from our ablations. A plain \textsc{Wer} reward suffices, more elaborate combinations do not help. No synthetic speech selection strategy consistently beats random sampling once the pool is moderately sized. And most of the achievable WER reduction comes from mixing 5 to 10\,h of real speech into a larger synthetic pool, rather than from annotating up to 54\,h in-domain.
Overall, when synthetic speech is the main or only adaptation resource for LLM-based ASR, reinforcement learning should be preferred over supervised fine-tuning.

\section*{Acknowledgements}
This work was supported by Idiap Research Institute and Uniphore collaboration project. Part of this work was also supported by EU Horizon 2020 project ELOQUENCE (grant number 101070558).

Specific sections were refined using Claude Flable 5 to ensure linguistic clarity. All experiments, results, and scientific claims are the authors' own.

\bibliographystyle{IEEEtran}
\bibliography{references}

\end{document}